\documentclass[10pt,twocolumn,letterpaper]{article}
\pdfoutput=1 
\usepackage{cvpr}
\usepackage{times}
\usepackage{epsfig}
\usepackage{graphicx}
\usepackage{amsmath}
\usepackage{amssymb}

\usepackage{rotating}
\usepackage[export]{adjustbox}
\usepackage{subcaption}
\makeatletter

\usepackage{booktabs}
\usepackage{array, makecell, tabularx}
\usepackage{multirow}

\newcommand{\mycaption}[2]{\caption{\textbf{#1.}\xspace#2}}

\usepackage[bottom]{footmisc}

\cvprfinalcopy 

\def\cvprPaperID{****} 

\begin{document}

\title{Improving Deep Stereo Network Generalization with Geometric Priors}

\author{Jialiang Wang\thanks{Work was done at NVIDIA}\\
Harvard University \\
\and
Varun Jampani\thanks{Current affliation: Google Research}\\
NVIDIA \\
\and
Deqing Sun\footnotemark[2]\\
NVIDIA \\
\and
Charles Loop\\
NVIDIA \\
\and
Stan Birchfield\\
NVIDIA \\
\and
Jan Kautz \\
NVIDIA
}

\maketitle

\begin{abstract}
End-to-end deep learning methods have advanced stereo vision in recent years and obtained excellent results when the training and test data are similar. However, large datasets of diverse real-world scenes with dense ground truth are difficult to obtain and currently not publicly available to the research community. As a result, many algorithms rely on small real-world datasets of similar scenes or synthetic datasets, but end-to-end algorithms trained on such datasets often generalize poorly to different images that arise in real-world applications.
As a step towards addressing this problem, we propose to incorporate prior knowledge of scene geometry into an end-to-end stereo network to help networks generalize better. 
For a given network, we explicitly add a gradient-domain smoothness prior and occlusion reasoning into the network training, while the architecture remains unchanged during inference.
Experimentally, we show consistent improvements if we train on synthetic datasets and test on the Middlebury (real images) dataset. 
Noticeably, we improve PSM-Net accuracy on Middlebury from 5.37 MAE to 3.21 MAE without sacrificing speed.
\end{abstract}

\section{Introduction}
\label{sec:intro}

Capturing scene geometry or depth is a basic task for computer vision. 
Despite the development of high-quality 3D sensors, there are still numerous drawbacks such as added hardware and cost. 
Stereo vision provides an excellent alternative where depth is estimated using two images captured simultaneously from two vantage points. 
Stereo estimation is a classical computer vision problem that has been intensively studied  due to its wide applications. 

Recent supervised deep neural networks have significantly improved the performance in stereo depth estimation. 
But these networks are data hungry, and it is difficult to obtain large and diverse real-world stereo depth datasets with dense and accurate ground truth. Existing datasets, e.g.,~\cite{Geiger2012KITTI,KITTI2015,MB2014,ETH3D}, rely on LiDAR or structured light to obtain the ground truth.
However, it is challenging to synchronize LiDAR with the stereo cameras, especially with moving objects. 
For example, creators of the KITTI stereo benchmark dataset~\cite{Geiger2012KITTI,KITTI2015} manually fit 3D models to cars to obtain depth ground-truth of moving pixels, and mask out bicyclists and pedestrians. 
Deep models trained with such datasets work well on test images of the same dataset but typically do not generalize well to other datasets. 
Since it is difficult to obtain a real-world dataset for diverse scene types, there is a necessity for the design of robust deep neural networks that can generalize well from training on synthetic data. 

\begin{figure*}[t!]
    \centering
    \includegraphics[width=\textwidth]{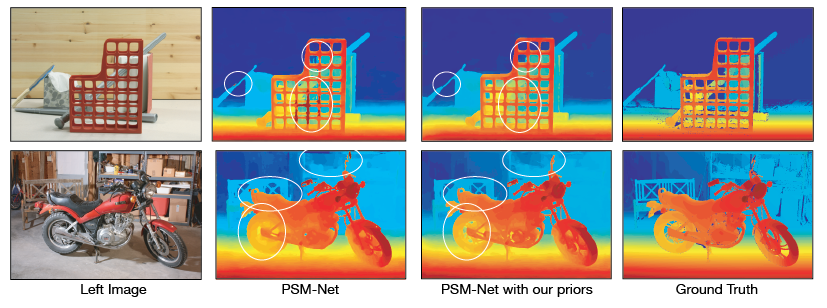}
    \mycaption{Overview}{We add geometric priors to an end-to-end stereo network to improve the robustness (generalization). The networks are trained with only synthetic images (FlyingThings3D~\cite{SceneFlow}) and tested with real images (Middlebury 2014~\cite{MB2014}). From left to right: (1) stereo left image; (2) baseline PSM-Net prediction; (3) Results with our proposed geometric priors to improve PSM-Net; (4) (almost-dense) ground truth disparity. Circles highlight regions with noticeable improvements. Zoom in to see details.}
    \label{fig:teaser_intro}
\end{figure*}

In this work, 
we propose techniques to improve generalization across datasets. We do this with novel training modules that incorporate scene geometry priors into the network training. Specifically, we propose two different training modules. The first module encourages piecewise smoothness in estimated depths. This is based on the common knowledge that scenes are usually composed of a number of continuous surfaces, and thus the depth varies piecewise smoothly across image pixels. The second module explicitly models the relationship between occlusions and disparities given by the rectified camera geometry. Together, these modules help the network capture geometric scene information that is invariant across different datasets. Since both network modules are back-propagatable, they can be used to augment any stereo network during training to improve the updating of the network parameters. During inference, however, only the original network is used, thus ensuring that neither runtime nor parameters are increased for any given network.


We empirically analyze the use of the proposed modules on two recent deep stereo networks: PSM-Net~\cite{PSM} and HSM-Net~\cite{HSM}. Experiments using synthetic training datasets (FlyingThings3D~\cite{SceneFlow} and Falling Things~\cite{FAT}) and real testing datasets (Middlebury~\cite{MB2014}) show consistent improvements due to our training modules. Figure~\ref{fig:teaser_intro} shows sample results indicating less noise in textureless regions and more fine details.

\section{Related Work}
\label{sec:related}

Since a comprehensive stereo review is out of scope,
we discuss the most relevant papers.

\vspace{1mm}
\noindent 
\textbf{Traditional approaches.} 
Classical methods typically first build a cost volume by comparing local patches at different disparity using image pixel intensity values or hand-crafted features (e.g.  census transform~\cite{zabih1994non}). The cost volume is then processed by a global optimization method, such as semi-global matching~\cite{hirschmuller2007sgm} or graph cut~\cite{kolmogorov2001computing} to get a regularized disparity map. Finally, some post-processing steps are usually adopted to improve the disparity map. 

\vspace{1mm}
\noindent \textbf{Deep-learning approaches.} Zbontar and LeCun~\cite{zbontar2016stereo} propose using Siamese networks to learn matching features to replace hand-crafted ones, and achieve better stereo results. Mayer \etal~\cite{SceneFlow} propose an end-to-end CNN stereo algorithm, which computes correlation between CNN features to construct a 1D cost volume and process it using CNN layers. Kendall \etal~\cite{kendall2017end} propose concatenating features to construct the cost volume and processing it using 3D convolutions, which is GPU-memory-intensive and relatively slow. 

Chang and Chen introduce the PSM-Net~\cite{PSM}, which uses spatial pyramid pooling modules and stacked-hourglass modules to improve the accuracy. However, PSM-Net still uses 3D convolutions and cannot process high-resolution images because of GPU memory constraints. Later methods reduce the computational overhead and/or aim for high-resolution images~\cite{khamis2018stereonet, HSM, tulyakov2018practical, wang2019anytime, tonioni2019real, yee2019fast}. In particular, HSM-Net~\cite{HSM} estimates disparity in a coarse-to-fine manner, and uses novel data augmentation methods to achieve state-of-the-art in mean absolute error (MAE) in the high-resolution Middlebury dataset while running with low latency. In this paper, we test our proposed priors on both PSM-Net and HSM-Net and show concrete improvements over the two strong baseline methods.

\vspace{1mm}
\noindent \textbf{Post-processing.}
Barron and Poole~\cite{cheng2018learning} propose using a fast bilateral solver on the disparity map predicted by the MC-CNN~\cite{zbontar2016stereo} and achieve a lower MAE. Cheng \etal~\cite{cheng2018learning} add a convolutional spatial propagation network module on top of a deep stereo architecture inspired by the PSM-Net. However, these works introduce additional computation at test time. In contrast, we add the proposed geometric priors during the training time to improve the parameters of the baseline networks without adding additional computation at test time.

\vspace{1mm}
\noindent \textbf{Prior constraints for stereo.} There are several notable works on improving stereo with prior constraints, e.g.~\cite{zhang2014rigid, woodford2009global}.
Most noticeably, Woodford \etal~\cite{woodford2009global} use visibility test and second-order smoothness priors in a classical stereo algorithm. They detect occlusions by the mapping uniqueness criterion~\cite{Brown2003advances,wei2005asymmetrical}. 
Our work revisits the second-order smoothness prior in an end-to-end trainable framework. We determine occlusion maps from disparity maps directly using the projective geometry (explained in Section \ref{sec:occ_disp}) and add a loss term that incorporates a gradient-domain prior to train a deep stereo  network. 

\begin{figure}[t!]
     \centering
     \begin{subfigure}[b]{0.32\textwidth}
         \centering
         \includegraphics[width=\textwidth]{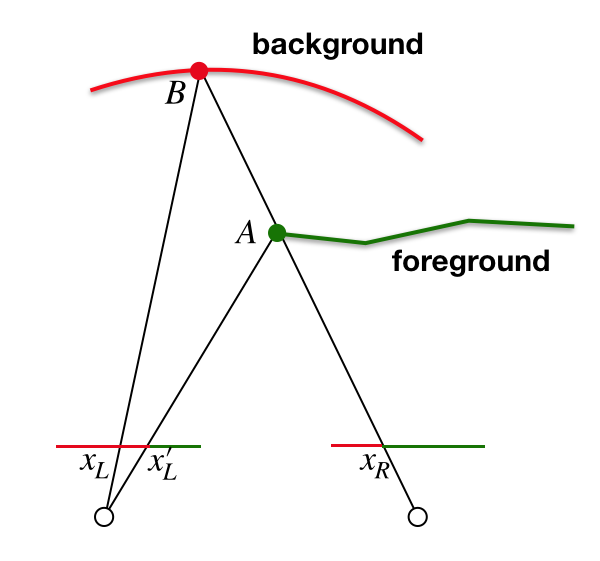}
         \caption{}
         \label{fig:y equals x}
     \end{subfigure}
     \hfill
     \begin{subfigure}[b]{0.4\textwidth}
         \centering
         \includegraphics[width=\textwidth]{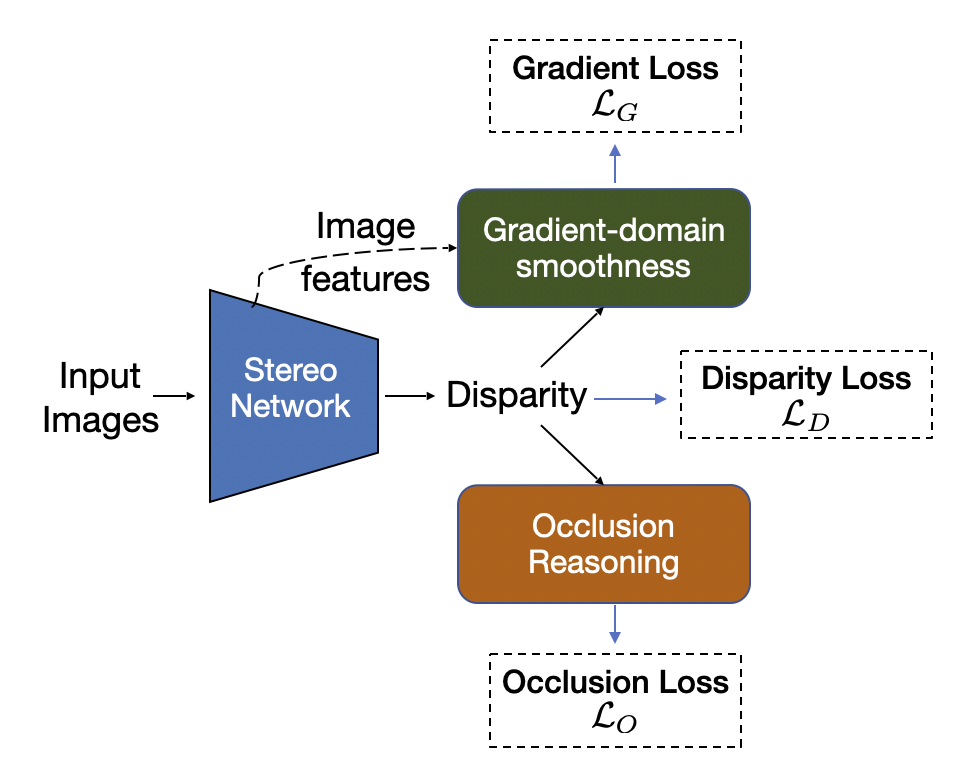}
         \caption{}
         \label{fig:three sin x}
     \end{subfigure}
     \caption{ \textbf{(a) Disparity and occlusion.} The disparity change at the occlusion boundary $D(x_A, y_A) - D(x_B, y_B)$, is equal to the pixel shift $x_L - x_L'$. \textbf{(b) Architecture overview.} We add the gradient-domain smoothness prior via PAC filtering and an occlusion reasoning module during the training of a stereo network. }
\label{fig:overview_and_occ}
\end{figure}

\section{Preliminaries}
In this section, we describe preliminaries of the two key components of our method: 1) pixel-adaptive convolutions, and 2) the relationship between disparities and occlusions for a rectified stereo image pair. 

\subsection{Pixel-adaptive convolutions}
\label{sec:pac}
We make use of recently proposed pixel-adaptive convolutions (PAC)~\cite{su2019pixel} 
to incorporate learned smoothness priors on estimated disparities. Here, we briefly review PAC and will describe the use of PACs in our framework in Section~\ref{sec:smoothness}. 
Following the notation from~\cite{su2019pixel}, standard convolution of image features
$\mathbf{v} = \{v_1, \cdots v_n \}$ with $c$ input channels and $c'$ output channels; $\mathbf{v}_i \in \mathbb{R}^{c}$
with filter $\mathbf{W}\in \mathbb{R}^{c' \times c \times s \times s}$ can be written as 
\begin{equation}
\mathbf{v}'_i = \sum_{j \in \Omega(i)} \mathbf{W}\left[\mathbf{p}_i - \mathbf{p}_j\right] \mathbf{v}_j + \mathbf{b},
\label{eq:conv}
\end{equation}
where $\mathbf{p}_i=(x_i,y_i)^\intercal$ are pixel coordinates,  $\mathbf{W}\left[\mathbf{p}_i - \mathbf{p}_j\right]$ denotes a 2D slice of the 4D tensor $\mathbf{W}$, $\Omega(\cdot)$ defines an $s\times s$ neighborhood, and $\mathbf{b}\in \mathbb{R}^{c'}$ denotes biases. One of the core properties of standard spatial convolution is spatial-invariance as the filter $\mathbf{W}$ only depends on position offsets $\left[\mathbf{p}_i - \mathbf{p}_j\right]$.
PAC provides a generalization of standard convolution in CNNs by adapting filter $\mathbf{W}$ at each pixel with a content-adaptive kernel $K$ that depends on pixel features $\mathbf{f}$:
\begin{equation}
\mathbf{v}'_i = \sum_{j \in \Omega(i)} K\left(\mathbf{f}_i, \mathbf{f}_j\right) \mathbf{W}\left[\mathbf{p}_i - \mathbf{p}_j\right] \mathbf{v}_j + \mathbf{b}, \label{eq:pac}
\end{equation}
where $K$ is a kernel function that has a fixed parametric form
such as Gaussian: $K(\mathbf{f}_i, \mathbf{f}_j)=\exp 
(-\frac{1}{2}(\mathbf{f}_i-\mathbf{f}_j)^\intercal 
(\mathbf{f}_i-\mathbf{f}_j))$, where $\mathbf{f}$ are \emph{guidance} features. PAC generalizes other widely used filtering operations such as bilateral filtering~\cite{aurich1995non, tomasi1998bilateral}. 

\subsection{Relationship between disparities and occlusions}
\label{sec:occ_disp}

Given any disparity map $D$, we can directly estimate an occlusion map $O$ in a principled way using rectified camera geometry~\cite{belhumeur1996bayesian, wang2019local}. Here we review the mathematical relationship using the left image as the reference image. Figure~\ref{fig:overview_and_occ}(a) shows an example of a foreground surface and a background surface. Assuming rectified geometry and both principal points located at the image center, by the definition of disparity we have
$D(x_A, y_A) = x_L' - x_R$
at the occlusion boundary point $A$ on the foreground, and
$D(x_B, y_B) = x_L - x_R$
at the first mutually visible pixel $B$ on the background. Subtracting both sides yields $D(x_A, y_A) - D(x_B, y_B) = x_L' - x_L$.
Therefore, the number of occluded pixels is equal to the disparity difference.


\begin{figure*}[t!]
    \centering
    \includegraphics[width=0.9\textwidth]{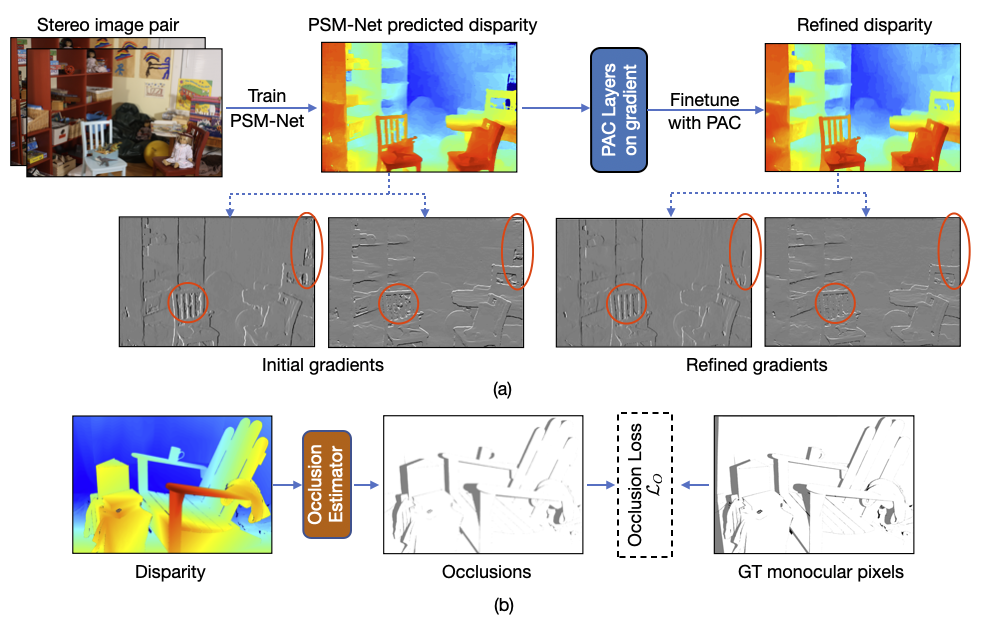}
    \mycaption{The effect of PAC and occlusion modules}{(a) We improve the initial disparity map by finetuning with a smoothness prior on the disparity gradient domain (PAC). The refined gradients are thin and less noisy, for example, at the chair and on the wall. This makes the disparity prediction more accurate. Circles highlight regions with noticeable improvements. (b) From a true disparity map (left), we get an estimated occlusion map (middle). For comparison, the Middlebury ground truth \textbf{monocular} map (right) is provided.} 
    \label{fig:pac_and_occ}
\end{figure*}

\section{Method}

We propose adding two priors into the training of an existing deep network as additional supervision.  Figure~\ref{fig:overview_and_occ}(b) shows the overview of our architecture. 

\subsection{Gradient-domain smoothness prior}
\label{sec:smoothness}
A typical stereo loss function minimizes some loss metric on disparity (e.g., $\mathcal{L}_1$ loss). This loss treats all regions on the image equally, and it sometimes fails to model desirable properties of the disparity.
Using the insight that many surfaces in the scene are approximately planar, we enforce a gradient-domain smoothness prior by processing the disparity gradient $D_x(x,y)$ and $D_y(x,y)$ with a gradient smoothness module consisting of PAC layers to filter the gradient to yield a refined gradient $\tilde{D}_x(x,y)$ and $\tilde{D}_y(x,y)$. 
Intuitively, PAC layers output similar 
values for two nearby pixels when the guidance features ($\mathbf{f}$ in Eqn.~\ref{eq:pac}) at those pixels are similar, while preserving the discontinuity at places where the features change dramatically (e.g., object boundaries).
Note that, unlike applying a PAC module directly to the disparity map $D(x,y)$ itself, which would tend to assign the same disparity values to surfaces with similar features, applying a PAC module to $D_x(x,y)$ and $D_y(x,y)$ (as we do) naturally models slanted planes and allows quadratic surfaces. Prior researchers have also found that second-order priors in the gradient domain can better model the world~\cite{woodford2009global, blake1987visual, grimson1981images, terzopoulos1983multilevel}.
Although the guidance features for the PAC module could simply be the RGBXY values of the image~\cite{su2019pixel}, additional information is captured by the image features that are already learned in early stages of the network. 
Experimentally we have found that, in many cases, the learned features achieve better performance. 

Figure \ref{fig:pac_and_occ}(a) shows an example of the effect of PAC layers on disparity gradients. The baseline PSM-Net~\cite{PSM} (top row) trained with only $\mathcal{L}_1$ disparity loss produces noisy disparity gradients in background slanted planes and near foreground object boundaries, whereas after finetuning with PAC-refined gradients loss, the noise is significantly reduced.

\begin{figure*}[t!]
    \centering
    \includegraphics[width=0.7\textwidth]{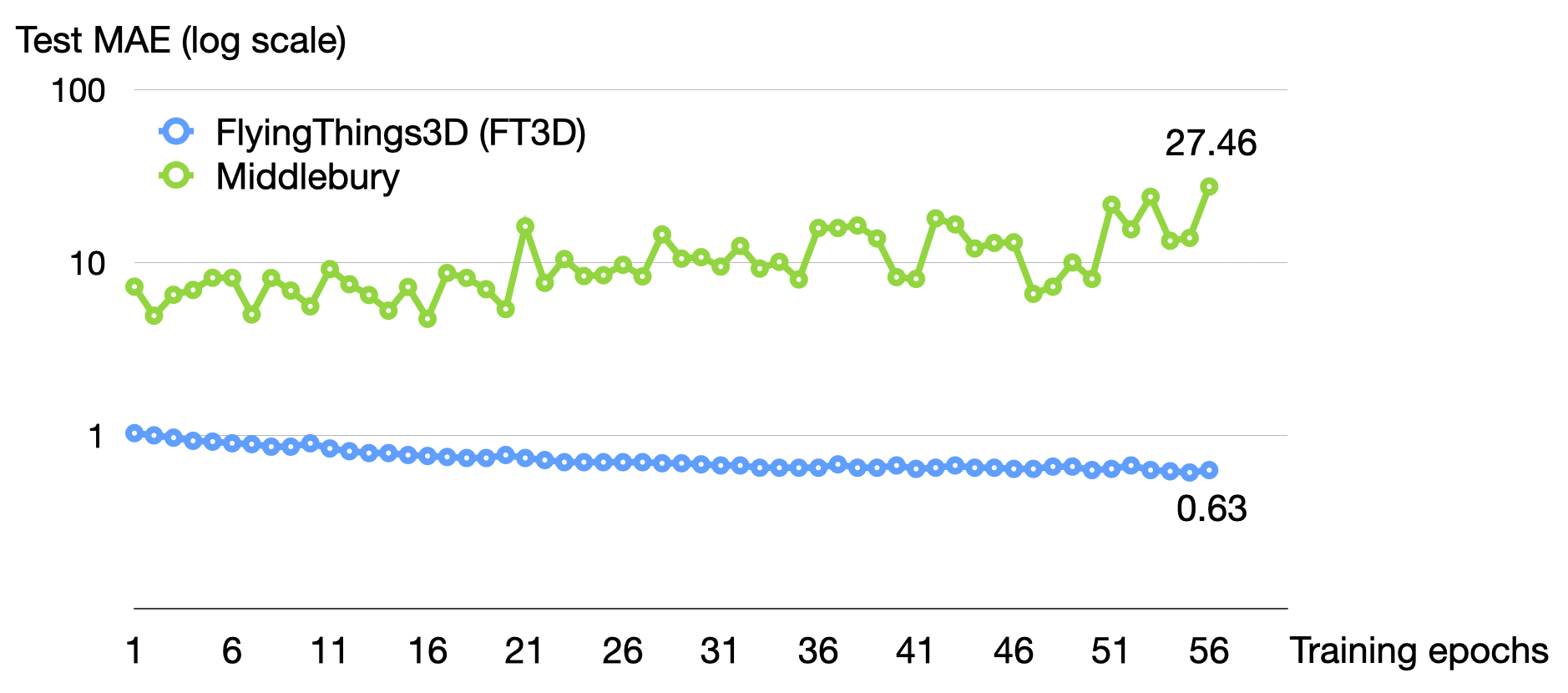}
    \caption{\textbf{Over-fitting.} MAE vs.~epochs for PSM-Net trained on FlyingThings3D. While more iterations improve the MAE on the FlyingThings3D test set, the MAE on Middlebury deteriorates rapidly.}
    \label{fig:PSM_overfit}
\end{figure*}

\subsection{Occlusion reasoning}
We can deterministically estimate
an occlusion map using the equations described in Section \ref{sec:occ_disp} without adding additional parameters into the baseline network. To compute a ``soft'' occlusion map $O$ at a pixel $(x,y)$ for back propagation, we use a ``steep'' Sigmoid function to approximate a step function:
\begin{equation}
\label{eqn:soft_occ}
    O(x,y) = \max_{x'>x} \biggl( \text{Sigmoid}(\alpha (\Delta d - \Delta x - d_0)) \biggl)
\end{equation}
\noindent where $\Delta d = D(x',y) - D(x,y)$ and $\Delta x = x' - x$. 
The parameter $\alpha$ controls the slope. The offset $d_0$ is needed because $\Delta x$ is always positive, leading to a skewed distribution.
We conducted a grid search to find the optimal parameters, and we found $\alpha = 3$ and $d_0 = 0.5$ gave us good occlusion $O(x,y)$ estimates. Figure \ref{fig:pac_and_occ}(b) shows an example of our estimated left occlusion map (middle) from the ground truth disparity map (left), in comparison with the ground truth left monocular mask (right) provided by the Middlebury dataset. 

\subsection{Total loss function}
We propose to finetune a state-of-the-art stereo network using the following loss function to improve the network's weights.
\begin{equation} 
\label{eq1}
\mathcal{L} = \mathcal{L}_D + \lambda_1 \mathcal{L}_G + \lambda_2 \mathcal{L}_O,  
\end{equation}
\text{where, }
\begin{equation*}
\begin{split}
    \mathcal{L}_D &= \mathcal{L}(D, D^{GT}) \\
    \mathcal{L}_G &= \mathcal{L}(\tilde{D}_x, D_x^{GT}) + \mathcal{L}(\tilde{D}_y, D_y^{GT}) \\
    \mathcal{L}_O &= \mathcal{L}(O, O^{GT})
\end{split}
\end{equation*}
\noindent Here, $\mathcal{L}_D$ is the disparity loss, $\mathcal{L}_G$ is the gradient domain smoothness loss and $\mathcal{L}_O$ is the occlusion loss. We use smooth $\mathcal{L}_1$ loss for $\mathcal{L}_D$, $\mathcal{L}_G$ and $\mathcal{L}_O$.

\begin{figure*}[t!]
    \centering
    \includegraphics[width=\textwidth]{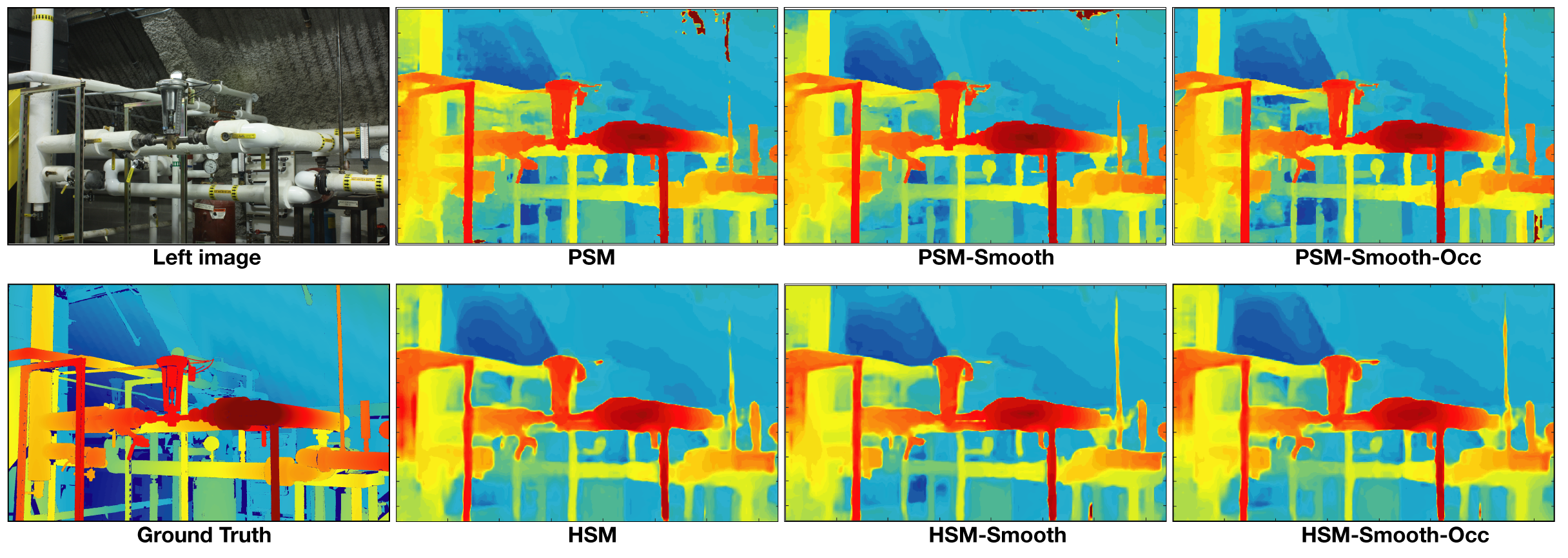}
    \mycaption{Generalization with geometric priors}{Models trained with our proposed geometric priors predict disparity maps with less noise on the real Middlebury dataset. We achieve the improvement without reducing speed in test time.}
    \label{fig:Mb_test_robustness}
\end{figure*}

\section{Experiments}
\label{sec:exps}
As discussed above, designing robust deep neural networks that can generalize well from training on synthetic data is especially important for stereo because the ground truth for real images is hard to obtain, and sometimes has its own noise (\eg the ``ground truth'' obtained using LiDAR). As a step towards this direction, we evaluate the proposed methods using three datasets: FlyingThings3D~\cite{SceneFlow} and Falling Things~\cite{FAT} as training datasets, and Middlebury~\cite{MB2014} as the test dataset. 

We analyze our proposed modules with two
state-of-the-art end-to-end trainable stereo networks discussed in Related Work, PSM-Net~\cite{PSM} and HSM-Net~\cite{HSM}.
Our technique is agnostic to the base stereo network architecture. That is,
we only add additional parameters and computations during training. So, our inference runtime is exactly the same as the baseline. 
For brevity, we will refer to our occlusion reasoning module as ``Occ'' and our gradient-smoothness module via PAC filtering as ``GradSmooth''.

\subsection{Datasets and Evaluation}

\noindent \textbf{FlyingThings3D}~\cite{SceneFlow} consists of a large number of synthetic images with 3D objects flying around in space.
It is one of the three scenes in the SceneFlow~\cite{SceneFlow} dataset. The FlyingThings3D test set is the same as the SceneFlow test set.

\vspace{1mm}
\noindent \textbf{Falling Things}~\cite{FAT} is a recent synthetic dataset of household objects with realistic rendering. We used the ``mixed'' sequences with more than one foreground object.

\begin{figure*}[t!]
    \centering
    \includegraphics[width=\textwidth]{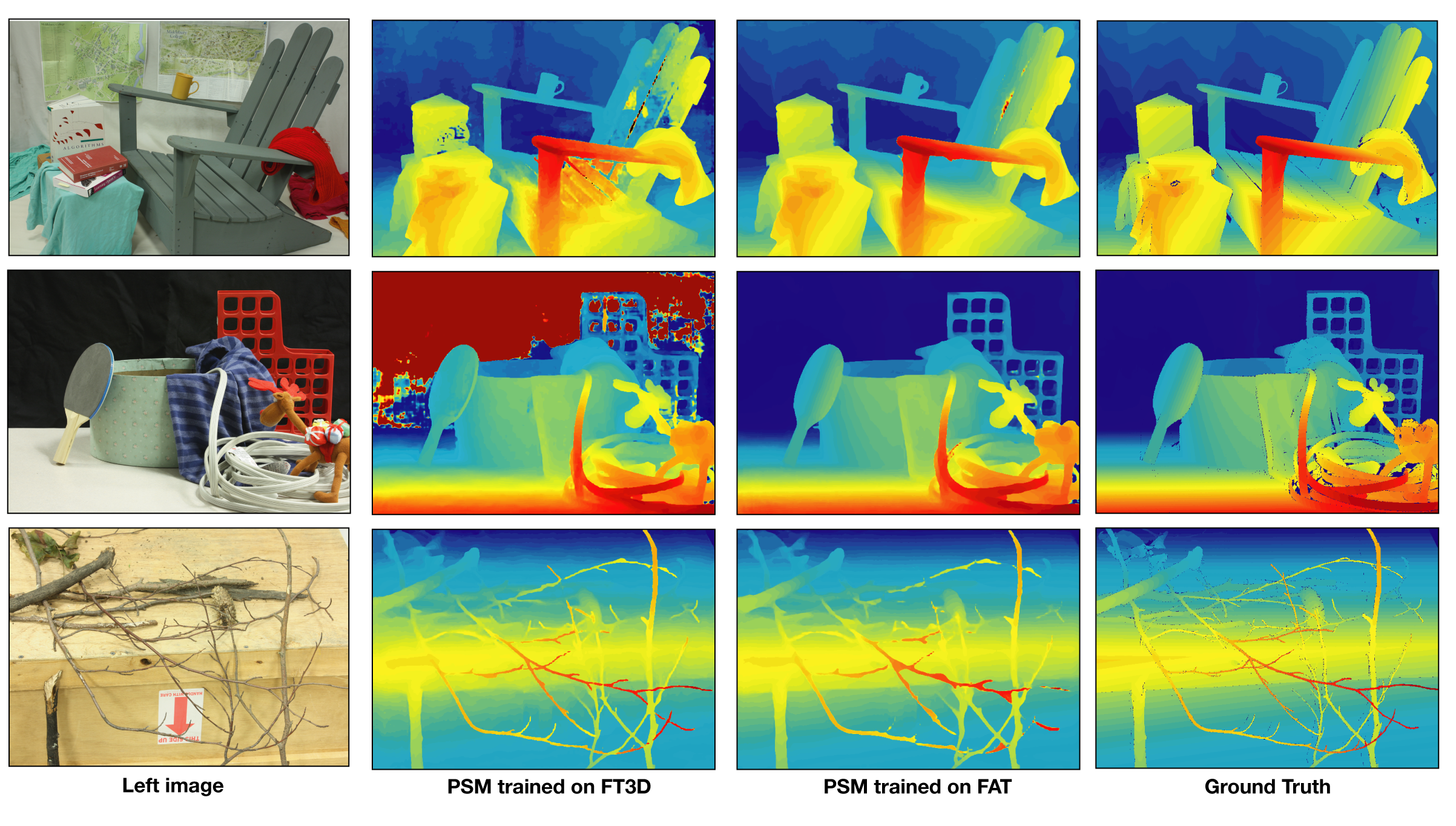}
    \mycaption{Generalization with different datasets}{PSM-Net~\cite{PSM} trained on Falling Things~\cite{FAT} improves the overall generalization in textureless or repeated-texture regions (Rows 1-2). However, in some cases, it fails to recover some very fine structures (row 3).}
    \label{fig:Mb_test_dataset}
\end{figure*}

\vspace{1mm}
\noindent \textbf{Middlebury 2014}~\cite{MB2014} is a high-resolution real-image stereo dataset. Figures \ref{fig:teaser_intro}, \ref{fig:pac_and_occ}, \ref{fig:Mb_test_robustness} and \ref{fig:Mb_test_dataset} 
show some dataset images. We test on all 23 Middlebury scenes that have ground truth, which include ``evaluation training sets''  and ``additional training images''. 
We use their ``perfect'' images with balanced lighting.  Since Middlebury images are of higher resolution than FlyingThings3D and Falling Things, we evaluate on quarter-resolution Middlebury images, similar to some other methods~\cite{hirschmuller2007sgm, PSM}.

\vspace{1mm}
\noindent \textbf{Evaluation Metrics}
All the results we report in this section are on \textbf{ALL} pixels that have ground truth. We evaluate using two standard stereo metrics: bad-2.0 (the percentage of pixels in which $\textbf{abs}(D^{Pred}(x,y) - D^{GT}(x,y)) > 2.0$, and MAE (the mean absolute error). 

\subsection{Implementation details}
In all our experiments, we use two PAC layers sequentially in a GradSmooth module. The first PAC layer is a $3\times 3$ convolutional layer with a dilation of 4. The second PAC layer is a $3\times 3$ convolutional layer with a dilation of 8 and then output a filtered disparity gradient. 
With PSM-Net, we add one GradSmooth module for each resolution, thus in total three GradSmooth modules. In HSM-Net, we add one GradSmooth module at the highest resolution . We set $\lambda_1 = \lambda_2 = 1.0$ in our loss function. In experiments without occlusion reasoning (denoted as PSM-GradSmooth or HSM-GradSmooth), we simply drop the occlusion term in the loss function. 
We use incremental training in our experiments, which means we first train a given base network, and then finetune it with our added modules.

\subsection{Robustness with geometric priors}
\label{sec:exp_robust}

\noindent \textbf{FlyingThings3D.} One of the main metrics researchers use to compare different stereo networks is the test set accuracy on the same (but disjoint) dataset images used in training (\textit{e.g.} SceneFlow).
We observe that if we finetune PSM-Net (starting with the authors' weights) with only FlyingThings3D data in the SceneFlow dataset, we can achieve state-of-the-art 0.63 MAE\footnote{Same as the original PSM-Net, we only evaluate on pixels with disparity $\le 192$ due to GPU memory limit.} for the SceneFlow test set. However, the network overfits to the synthetic data, as evidenced by the very high test error we observe on Middlebury (27.46 MAE). See Figure \ref{fig:PSM_overfit}. This suggests that  network robustness is a more important metric to optimize. Specifically, we define robustness as the network's ability to generalize when trained on synthetic images and then tested on real images. As discussed before, it is highly desirable to design and train networks using synthetic images that can generalize well.

Table~\ref{tab:robustness_ft3d} shows the quantitative comparison. For PSM-Net, with the author's published weights trained on SceneFlow~\cite{PSM}, the test MAE on Middlebury is 16.91. After finetuning only on FlyingThings3D, the best model's MAE is reduced to 5.37. After further finetuning with the GradSmooth module, the MAE is further reduced to 4.62, and with both GradSmooth and Occ modules, the error is reduced to 3.21. Again, this is all done without reducing speed at test time.

\begin{table}[h]
\centering
\small
\begin{tabular}{l c c c c}
\toprule
Architecture & Train  & Test & bad-2.0 & MAE \\ 
\midrule
PSM (author weights) & SF & MB & 29.67 & 16.91\\ 
PSM & FT3D & MB & 19.58 & 5.37\\ 
PSM-GradSmooth& FT3D & MB  & 19.04 & 4.62 \\ 
PSM-GradSmooth-Occ & FT3D & MB  & \textbf{18.17} & \textbf{3.21} \\ 
\midrule
HSM & FT3D & MB  & 25.56 & 3.00 \\ 
HSM-GradSmooth & FT3D & MB  & \textbf{22.09} & \textbf{2.71} \\ 
HSM-GradSmooth-Occ & FT3D & MB  & 25.24 & 3.06 \\ 
\bottomrule
\end{tabular}
\mycaption{Robustness with geometry priors}{The smoothness prior improves generalization in both PSM-Net and HSM-Net whereas the occlusion prior further improves PSM-Net. MB: Middlebury, FT3D: FlyingThings3D. SF: SceneFlow} 
\label{tab:robustness_ft3d}
\end{table}

For HSM-Net~\cite{HSM}, we observe that the GradSmooth module helps improve upon the baseline HSM-Net, but not the Occ module. One possible reason could be that the sophisticated occlusion augmentation that HSM-Net performs on the training images may have a complicated interaction with our occlusion module. Further, the ground truth disparity may be missing in some occluded regions on the Middlebury dataset, making it hard to evaluate the improvement brought by the occlusion reasoning module.

Figures~\ref{fig:teaser_intro} and~\ref{fig:Mb_test_robustness} show some qualitative results. In general, networks trained with our geometric priors predict disparity maps with less noise in the smooth regions, and the boundaries align better with the actual object boundaries than those trained without the geometric priors.

\vspace{1mm}
\noindent \textbf{Falling Things. } We also train PSM-Net with the newer Falling Things dataset \cite{FAT} and compare the results with models trained with the more popular FlyingThings3D dataset. Table \ref{tab:robustness_dataset} shows that PSM-Net trained with Falling Things significantly outperforms the same network trained on FlyingThings3D \cite{SceneFlow}. Even with Falling Things, the GradSmooth module we propose still helps the performance, improving the percentage of bad pixels to 10.35.

Figure \ref{fig:Mb_test_dataset} shows the qualitative comparison between PSM-Net trained on FlyingThings3D and Falling Things. PSM-Net trained with FlyingThings3D often fails on textureless regions (rows 1-2). However, the predictions by models trained on Falling Things sometimes lack fine details (\textit{e.g.} row 3). 

\begin{table}[h]
\centering
\begin{tabular}{l c c c c}
\toprule
Architecture & Train  & Test & bad-2.0 & MAE \\ 
\midrule
PSM & FT3D & MB & 19.58 & 5.37\\ 
PSM & FT3D & MB & 11.48 & \textbf{1.41} \\ 
PSM-GradSmooth & FT3D & MB & \textbf{10.35} & 1.42   \\ 
\bottomrule
\end{tabular}
\mycaption{Dataset effect}{Training PSM-Net on Falling Things~\cite{FAT} improves the accuracy. Our proposed second-order smoothness prior helps further in percentage of bad pixel error. MB: Middlebury, FT3D: FlyingThings3D.}
\label{tab:robustness_dataset}
\end{table}

\section{Conclusion}
\label{sec:conclusion}
As a step towards designing robust stereo networks, we propose adding two priors into the stereo networks based on our knowledge of scene geometry. Our proposed geometric priors neither add computation nor require extra memory at inference time.  Extensive experiments show that adding a gradient-domain smoothness prior consistently improves generalization and, in some cases, adding occlusion prior further improves generalization. 

\clearpage

{\small
\bibliographystyle{ieee_fullname}
\bibliography{egbib}
}

\end{document}